%% file: root.tex
\documentclass[letterpaper, 10 pt, journal, twoside]{IEEEtran}  % 

\IEEEoverridecommandlockouts                              % This command is only needed if
\makeatletter
\let\NAT@parse\undefined
\makeatother
\usepackage[numbers,sort&compress]{natbib}

\usepackage{graphicx}
\usepackage{amsmath}
\usepackage{amssymb}  % assumes amsmath package installed
\usepackage{subcaption}
\usepackage{multirow}
\usepackage{array,booktabs}
\usepackage{diagbox}
\usepackage{balance}
\usepackage{xspace}
\usepackage{algorithm}
\usepackage{algpseudocode}
\usepackage{adjustbox}
\usepackage{mathtools}

\usepackage[final]{hyperref}
\hypersetup{
 colorlinks=true,
 linkcolor=magenta,
 filecolor=magenta,
 urlcolor=magenta,
 citecolor=magenta
}
\usepackage{cleveref}
\crefname{figure}{Fig.}{Figs.}
\Crefname{figure}{Fig.}{Figs.}
\usepackage[font=small]{caption}
\usepackage{rpm_SIunits}
\usepackage{rpm_acronyms}
\usepackage{rpm_math}
\usepackage{rpm_misc}
\usepackage{bbm}

\usepackage{textcomp}

\usepackage{soul,color}
\usepackage{lipsum}
\usepackage{dsfont}

\usepackage[table]{xcolor}
\usepackage{colortbl}

\newcommand{\eg}{\textit{e.g.}}
\newcommand{\ie}{\textit{i.e.}}

\usepackage{amssymb}% http://ctan.org/pkg/amssymb
\usepackage{pifont}% http://ctan.org/pkg/pifont
\newcommand{\cmark}{\ding{51}}%
\newcommand{\xmark}{\ding{55}}%
\usepackage{tikz} % \checkmark

\title{\LARGE \bf TRaIL-Odom: Tightly Coupled Continuous Time Radar-IMU-LiDAR Odometry with Adaptive Doppler Weighting
}     

\author{Chiyun Noh${}^{1}$, Turcan Tuna${}^{2}$, William Talbot${}^{2}$, Marco Hutter${}^{2}$, Laurent Kneip${}^{3*}$, and Ayoung Kim${}^{1*}$
\thanks{Manuscript received: April 21, 2026; Revised July 12, 2026; Accepted August 23, 2026.}%Use only for final RAL version
\thanks{This paper was recommended for publication by Editor Javier Civera upon evaluation of the Associate Editor and Reviewers’ comments.
This work was supported by the NRF grant (No. RS-2023-00241758) and KIAT (P0020536), and in part by the Robotics and AI (RAI) Institute. (\textit{Corresponding authors: Laurent Kneip and Ayoung Kim.)}}
\thanks{$^{1}$C. Noh and A. Kim are with the Department of Mechanical Engineering, Seoul National University, S. Korea.
        {\tt\footnotesize [gch06208, ayoungk]@snu.ac.kr}}%
\thanks{$^{2}$T. Tuna, W. Talbot and M. Hutter are with the
Robotic Systems Lab (RSL), ETH Z\"urich, 8092 Z\"urich, Switzerland. 
        {\tt\footnotesize [tutuna, wtalbot, mahutter]@ethz.ch}}%
\thanks{$^{3}$L. Kneip is with the Robotics and AI Institute (RAI), Z\"urich, Switzerland. {\tt\footnotesize lkneip@rai-inst.com}}
\thanks{Digital Object Identifier (DOI): see top of this page.}
}

\begin{document}

%\onecolumn
\maketitle

\input{sections/0_abstract}

\begin{IEEEkeywords}
SLAM, Localization, Range Sensing
\end{IEEEkeywords}

\input{sections/1_introduction}

\input{sections/2_relatedwork}

\input{sections/3_method}
\input{sections/4_experiment}
\input{sections/5_conclusion}

% \newpage
% \newpage

%\section*{ACKNOWLEDGMENT}
% \balance
\small
\bibliographystyle{IEEEtranN} %citeauthor
\bibliography{string-short,references}

\end{document}

%% file: sections/0_abstract.tex
\begin{abstract}
Existing radar-LiDAR fusion methods rely on fixed residual weights, even though the informativeness of radar Doppler and LiDAR geometry is scan- and direction-dependent, leading to uniform radar weighting that misallocates Doppler information across translational directions.
To address this limitation, we propose two degeneracy-aware Doppler reweighting modules within a tightly coupled Radar-IMU-LiDAR odometry framework: per-point radar reweighting and scan-wise radar gain scheduling. Since geometric degeneracy is directional, we first identify weak translational directions from the LiDAR geometry and reweight individual radar Doppler constraints based on their alignment with the weak subspace. We further adjust the overall radar contribution using LiDAR geometric anisotropy such that radar is emphasized when LiDAR observability is poor and suppressed when LiDAR constraints are already reliable. Across 13 evaluated sequences, TRaIL-Odom achieves state-of-the-art overall performance, with clear advantages in geometrically degenerate scenes. In ablation experiments on three degenerate sequences, combining the two adaptive weighting modules reduces RMSE ATE and RTE by 86.0\% and 78.5\% relative to the fixed-weight baseline. We make our code and an accompanying dataset publicly available at
\href{https://github.com/ChiyunNoh/TRaIL-Odom}{https://github.com/ChiyunNoh/TRaIL-Odom}.
\end{abstract}

%% file: sections/1_introduction.tex
\section{Introduction}
\label{sec:intro}

\IEEEPARstart{R}{adar} has emerged as a promising complementary sensing modality due to its ability to provide direct radial velocity measurements. Building on these advantages, prior work~\cite{cynoh2025garlio, nissov2024Degrad, nissov2024Robust, hatleskog2025} incorporate radar Doppler information into \ac{LIO} to improve robustness in geometrically uninformative environments where LiDAR constraints become weak. However, most of these approaches still rely on fixed residual weights based on sensor noise models, leaving open the question of how radar contribution should be modulated under scene-dependent geometric degradation.

Rigid frameworks often require substantial tuning and show limited generalization across varied scenarios, because simplifying assumptions (\eg, Gaussian measurement noise and redundancy averaging) can break down under degenerate conditions, and corrupted measurements may dominate certain state directions~\cite{zhao2025resilient}. Ideally, if the true scene-dependent uncertainty of each residual block were known, uncertainty-based weighting would provide a principled way to balance modalities. In practice, this information is not available \emph{a priori}, motivating adaptive reweighting driven by measurable signals such as scan geometry and observability.

More importantly, geometric degeneracy is often directional rather than uniform across all directions~\cite{tunaInformed}. This means that, even within a scan, only a subset of translational directions are weakly constrained, while the remaining directions are already supported by LiDAR geometry. This suggests that radar should not be weighted uniformly. It should primarily reinforce weakly constrained directions, while its influence should be limited along directions that are already well constrained by LiDAR. This calls for an adaptive mechanism that modulates radar contribution both per measurement and per scan, rather than applying a single fixed weight throughout the trajectory.

\begin{figure}[!t]
  \centering
  \includegraphics[width=\columnwidth]{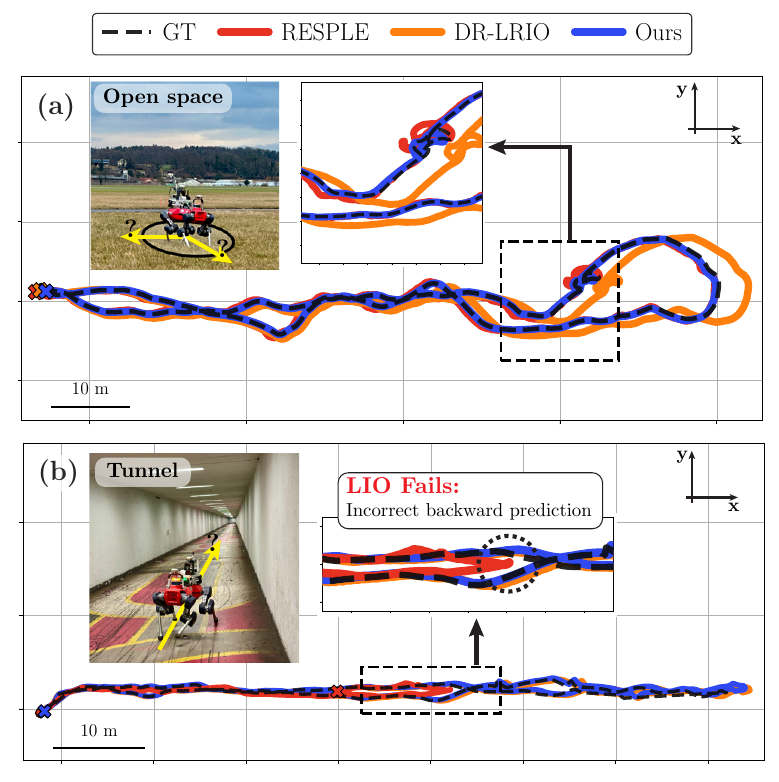}
  \caption{Results of \texttt{RESPLE} (LIO), \texttt{DR-LRIO} (LIO + radar), and \texttt{Ours} in degenerate environments: (a) \textit{Airfield1} and (b) \textit{BikeTunnel2}. While \texttt{RESPLE} is unstable in \textit{Airfield1} and fails in \textit{BikeTunnel2}, \texttt{Ours} remains stable in both. Zoomed views and inset images show local details and environments. Circles and crosses denote the start and end points, which overlap in the return-to-start trajectories.}
  \label{fig:Teaser}
  \vspace{-7mm}
\end{figure}

To address this, we propose \textbf{TRaIL-Odom}, a tightly coupled Radar-IMU-LiDAR optimization framework based on continuous time B-splines for fusing asynchronous measurements. First, for each LiDAR scan, geometric information is analyzed to identify a weak translational subspace. Then, constraints from radar Doppler measurements are reweighted to reinforce this subspace, without unnecessarily emphasizing radar along directions already well constrained by LiDAR. In addition, a scan-level radar gain is scheduled based on the severity of geometric degeneracy to adjust the overall Doppler contribution, improving stability and limiting unnecessary influence in well-constrained scenes.

The main contributions of this work are as follows:
\begin{itemize}
    \item We present TRaIL-Odom, a tightly coupled Radar-IMU-LiDAR optimization-based odometry approach using continuous time B-splines, designed for diverse real-world environments.

    \item We propose a degeneracy-aware per-point radar reweighting scheme that reallocates Doppler contribution toward weak translational directions identified from pointcloud geometry, while avoiding unnecessary radar influence along directions already well constrained.

    \item We further introduce a scan-wise radar gain scheduling strategy that adjusts the Doppler contribution according to LiDAR pointcloud anisotropy, emphasizing radar influence in scans with severe degeneracy while limiting unnecessary influence in well-constrained scans.
    
    \item We release a real-world dataset comprising 6 sequences across three scenes with degenerate scan geometries and accurate position ground truth. The method is validated on this dataset with extensive ablation studies. The code is also released to the research community.
\end{itemize}

%% file: sections/2_relatedwork.tex
\section{related work}
\label{sec:relatedwork}

\subsection{Radar-LiDAR Fusion}
Radar provides \ac{LOS}-direction Doppler measurements that complement LiDAR geometry, making Radar-LiDAR fusion effective in perceptually degraded or geometrically degenerate scenes. Prior work by ~\citet{nissov2024Degrad, nissov2024Robust} present a tightly coupled sliding-window estimator that jointly optimizes radial-velocity factors with LiDAR residuals, improving robustness under weak geometric constraints. In parallel, filter-based methods~\cite{cynoh2025garlio, qian2025af-rlio} integrate radar with \ac{LIO} and highlight additional capabilities such as velocity-aware gravity estimation and dynamic object removal. Beyond robustness-oriented designs, computational aspects of asynchronicity have also been addressed. ~\citet{hatleskog2025} propose IMU-preintegrated radar velocity factors to avoid adding states at every radar timestamp, reducing the number of state nodes and runtime in fixed-lag smoothing. 

However, many existing works are often validated on limited scenarios, and explicit modulation of radar contribution remains underexplored. To promote robust generalization, this work modulates the Doppler contribution according to LiDAR geometric observability, directing kinematic information toward weak directions while suppressing unnecessary influence in well-constrained scenes.

\subsection{Adaptive Fusion and Reweighting Under Degradation}
Robust operation in degraded scenes often relies on complementary modalities beyond geometry, such as LiDAR intensity or vision~\cite{coinlio2024,khedekar2025pg,zheng2024fastlivo2}. However, adding modalities alone does not guarantee higher accuracy, since relative measurement uncertainty across modalities is scene-dependent under degradation, and fixed fusion weights can over-trust unreliable measurements. 
Accordingly, several methods explicitly modulate the relative influence of different information sources, adapting each contribution as conditions change. ~\citet{zhao2025resilient} proposes a hierarchical adaptation framework that assigns direction-dependent feature weights based on degradation severity. Similarly,~\citet{du2025good} adaptively adjusts dead-reckoning priors in visual SLAM, and ~\citet{yun2025more} down-weight measurements in multi-camera RGB-D inertial odometry using photometric confidence. Even within a single modality, GenZ-ICP~\cite{lee2024genz} adapts complementary geometric error terms based on surrounding geometry to improve robustness in corridor-like degeneracy.

 Our approach shares the goal of adaptation under direction-dependent degradation but targets the reweighting of kinematic information from radar Doppler within a tightly coupled Radar-IMU-LiDAR optimization.
 Unlike methods that switch modules or globally adjust parameters, we use LiDAR scan geometry to identify weak translational directions and then adapt Doppler residual contributions per measurement and per scan to reinforce those directions while limiting influence when LiDAR constraints are already informative.

%% file: sections/3_method.tex
\section{Method}

\subsection{Overview}
\begin{figure}[!t]
  \centering
  \includegraphics[width=0.9\columnwidth]{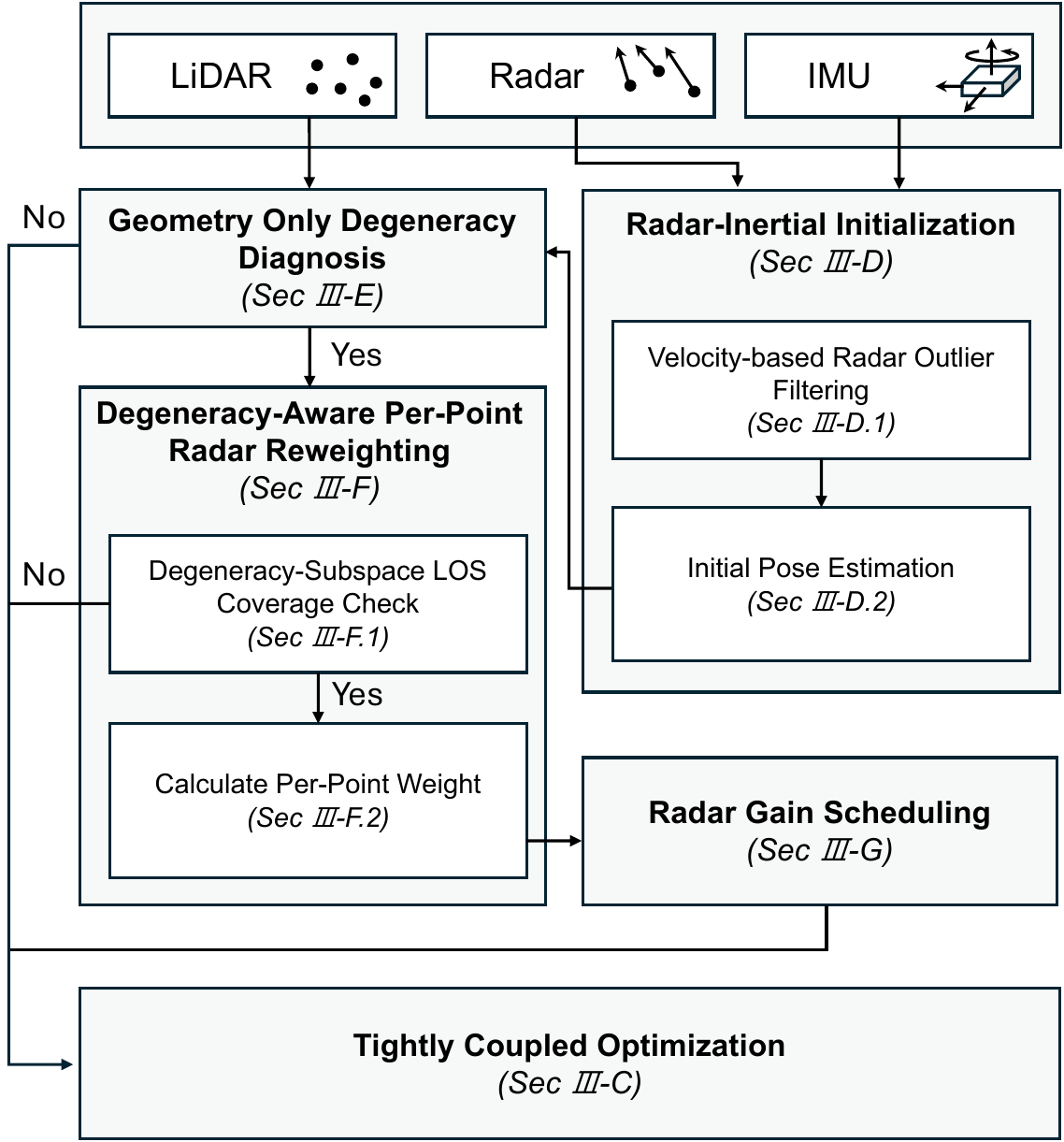}
  \caption{\textbf{Overview of TRaIL-Odom}. For each LiDAR scan, radar point clouds are filtered by velocity using the previous window estimate, followed by \ac{RI} initialization. LiDAR geometry is then used to diagnose translational degeneracy and reweight radar Doppler constraints. The weighted residuals are jointly minimized in a tightly coupled optimization to estimate pose.}
  \label{fig:Overview}
  \vspace{-7mm}
\end{figure}

We present a tightly coupled Radar-IMU-LiDAR optimization framework that leverages continuous time B-splines to fuse asynchronous measurements across diverse environments, with particular emphasis on geometrically degenerate scenes. The pose is modeled by separate position and orientation splines, known as the \emph{split pose} representation~\cite{talbot2025continuous}, which has computational advantages over a single $\mathrm{SE(3)}$ spline.
Our framework is summarized in \figref{fig:Overview}.

\subsection{Notation}
We follow the notation convention in \cite{slam-handbook}. Scalars are denoted by lowercase letters (\eg, $a$), column vectors by bold lowercase letters (\eg, $\boldsymbol{a}$), and matrices by bold uppercase letters (\eg, $\boldsymbol{A}$). A vector expressed in frame $\mathcal{F}^{a}$ is rotated to frame $\mathcal{F}^{b}$ using ${\boldsymbol{R}}^{b}_{a} \in \mathrm{SO}(3)$. The vector $\boldsymbol{t}{}^{b}_{a} \in \mathbb{R}^3$ denotes the position of the origin of frame $\mathcal{F}^{a}$ expressed in $\mathcal{F}^{b}$. Measured quantities are indicated by a tilde $\tilde{(\cdot)}$.
For state representation, we parameterize the continuous time body-to-world transformation using cubic uniform B-splines. We set the body frame to the IMU frame ($\mathcal{F}^{I}$) and initialize the world frame to the $\mathcal{F}^{I}$ at $t_0$. The trajectory is represented by a translational spline $\boldsymbol{p}(t) \in \mathbb{R}^3$ and a rotational spline $\boldsymbol{R}(t) \in \mathrm{SO}(3)$. Each spline is defined by a sequence of control points with uniform spacing $\Delta t$. In our implementation, we set $\Delta t=\unit{0.02}{\second}$. We adopt the cumulative formulation of \citet{sommer2020efficient}, which enables the efficient computation of temporal derivatives~\cite{talbot2025continuous}. The state is written as
\begin{eqnarray} 
\label{eq:state}
    \boldsymbol{x}(t) &\triangleq&
    \begin{bmatrix}
    \boldsymbol{R}(t) &
    \boldsymbol{p}(t) &
    \boldsymbol{b}_{g} &
    \boldsymbol{b}_{a}
    \end{bmatrix}.
\end{eqnarray}
The gyroscope and accelerometer biases of the IMU are $\boldsymbol{b}_g \in \mathbb{R}^3$ and $\boldsymbol{b}_a \in \mathbb{R}^3$ respectively. We assume that these biases remain constant over the duration of a single LiDAR scan. The gravity vector $\mathbf{{g}}^\mathtt{G}$ is initialized at startup from a short accelerometer average and kept fixed thereafter.

\subsection{Tightly Coupled Optimization}
\label{sec:joint_optimization}
State estimation is formulated as a single tightly coupled nonlinear least squares problem over a sliding window.
The window spans one LiDAR scan and includes the IMU and radar measurements collected during the interval. The optimization variables consist of the active B-spline control points parameterizing $\boldsymbol{R}(t)$ and $\boldsymbol{p}(t)$, together with the IMU biases.
The cost function and residual terms are defined as

\small
\begin{equation}
\label{eq:tightly-coupled}
{
\begin{aligned}
F(\boldsymbol{x}) =& \sum_{k \in \mathcal{I}}
    \Bigl(
    w_\omega \left\|\mathbf{r_{\omega_k}}(t_k)\right\|_2^2
    + w_a \left\|\mathbf{r_{a_k}}(t_k)\right\|_2^2
    \Bigr)\\
&+\sum_{k \in \mathcal{R}_{\mathrm{in}}} {w_{R,k}^{\textbf{*}}} \left\|\mathbf{r_{R_k}}(t_k)\right\|_2^2
+\sum_{k \in \mathcal{L}} w_L \left\| \mathbf{r_{L_k}}(t_k) \right\|_2^2\\
&+{{w_p}} {\left\| {{{\mathbf{r_p}}}}\right\|_2^2},
\end{aligned}}
\end{equation} 

\begin{equation}
\begin{aligned}
  \label{eq:lidar_residuals}
\mathbf{r_{L_k}}(t_k) =& \;\boldsymbol{n}_k^\top(\boldsymbol{R}(t_k){\boldsymbol{R}}^{{I}}_{L}\;\tilde{\boldsymbol{p}}_k+\boldsymbol{p}(t_k)-\boldsymbol{q}_k),
\end{aligned}
\end{equation}

\begin{equation}
\begin{aligned}
  \label{eq:imu_residuals}
  \mathbf{r_{\omega_k}}(t_k) &= \boldsymbol{\omega}(t_k)  +\boldsymbol{b}_{g}-\tilde{\boldsymbol{\omega}}_{m_k}
    \\
    \mathbf{r_{a_k}}(t_k) &= \boldsymbol{R}(t_k)^{\top}(\ddot{\boldsymbol{p}}(t_k) - \mathbf{{g}}^\mathtt{G})+\boldsymbol{b}_{a} - \tilde{\boldsymbol{a}}_{m_k},
\end{aligned}
\end{equation}

\begin{equation}
\begin{aligned}
  \label{eq:Radar_residuals}
    \mathbf{r_{R_k}}({t_k})= - \frac{(\tilde{\boldsymbol{p}}^{R}_k)^\top{}}{\left\| \tilde{\boldsymbol{p}}^{R}_k\right\|} \left( {\boldsymbol{R}}^{I}_{R} \right)^{\top}\left({\boldsymbol{R}(t_k)}^\top{\boldsymbol{\dot{p}}(t_k)}+\lfloor \boldsymbol{\omega}(t_k) \rfloor_{\times} \boldsymbol{t}{}^{I}_{R}\right) \\ - \tilde{v}^{R}_{m_k}
\end{aligned}
\end{equation}
\normalsize
where $\mathcal{I}$, $\mathcal{R}$, and $\mathcal{L}$ denote the IMU, radar Doppler, and LiDAR measurements in the window, respectively, $\mathbf{r}_p$ is the marginalization prior from the previous window computed using the Schur complement, $w_*$ denotes the scalar weight for each residual term, $\boldsymbol{\omega}(t)$ denotes the body angular velocity at time $t$, and $\lfloor \cdot \rfloor_\times$ denotes the skew-symmetric matrix operator. The radar Doppler weight $w_{R,k}^{\textbf{*}}$ is the effective Doppler weight, defined in Sec.~\ref{sec:Radar_gain_scheduling}. The residuals $\mathbf{r_{\omega_k}}(\cdot)$, $\mathbf{r_{a_k}}(\cdot)$, $\mathbf{r_{R_k}}(\cdot)$, and $\mathbf{r_{L_k}}(\cdot)$ denote the IMU gyroscope, IMU accelerometer, radar Doppler, and LiDAR point-to-plane terms, where $\boldsymbol{n}_k$ and $\boldsymbol{q}_k$ in \eqref{eq:lidar_residuals} denote the unit normal and a reference point of the associated plane in the local map.

In our implementation, the residual weights were empirically selected as nominal residual-scaling coefficients for stable optimization. A single global set was used across all reported experiments without sequence-specific tuning.

% Equation~\eqref{eq:tightly-coupled} defines a nominal fixed weight least squares objective.
% In degraded scenes, observability can collapse along particular state directions, and fixed weights do not actively reinforce those weak directions, which can leave the update poorly conditioned and the estimate unstable \cite{zhao2025resilient}.
% We therefore adapt the radar Doppler contribution per scan and per measurement using LiDAR diagnosed weak translational directions (Sec.~\ref{sec:Radar_reweighting} to \ref{sec:Radar_gain_scheduling}).

Equation~\eqref{eq:tightly-coupled} defines the tightly coupled least-squares objective. In degraded scenes, LiDAR geometric constraints can become weak or anisotropic along particular translational directions, making those directions poorly observable. Fixed weights alone cannot actively reinforce these LiDAR-weak directions, motivating the use of complementary sensing information to support them. We therefore adaptively modulate the nominal radar Doppler weight through the per-point factor $\alpha_k$ and the scan-wise gain $\gamma$ to complement LiDAR-diagnosed weak translational directions (Sec.~\ref{sec:Radar_reweighting} to \ref{sec:Radar_gain_scheduling}).

\subsection{\acf{RI} Initialization}
\label{sec:initialization_rio}
The objective in \eqref{eq:tightly-coupled} is solved iteratively and requires a reliable initialization for each window. We therefore use \ac{RI} initialization to provide a velocity-consistent starting point for the joint optimization, which has shown stable behavior in prior work~\cite{noh2025garlileo} even without LiDAR geometric information.

\subsubsection{Velocity-based Radar Outlier Filtering}
\label{Velocity-based Radar Outlier Filtering}
Radar measurements are prone to outliers arising from dynamic objects and multipath effects. Accordingly, radar measurements are first filtered prior to the \ac{RI} optimization. Outliers are suppressed via velocity-based filtering using the ego velocity $\boldsymbol{v}_{prior}^{R}$ in the radar frame, predicted from the previous optimization window at the timestamp of its last radar measurements. Specifically, for each detection, the predicted radial component is computed by projecting $\boldsymbol{v}_{prior}^{R}$ onto the point \ac{LOS}, and measurements whose radial velocities $\tilde{v}_{m_j}^{R}$ deviate significantly from this prediction are discarded as follows:

\small
\begin{equation}
\label{eq:velocity_filtering}
\mathcal{R}_{\mathrm{in}}
=\left\{\, j \in \mathcal{R} \; \middle|\ \bigl|-\tilde{v}_{m_j}^{R}-\tilde{\boldsymbol{l}}_j^{R\top}\boldsymbol{v}_{prior}^{R} \bigr|\le\delta_v \right\}.
\end{equation} \normalsize
where $\tilde{\boldsymbol{l}}_j^{R}\triangleq \tilde{\boldsymbol{p}}_j^{R}/\|\tilde{\boldsymbol{p}}_j^{R}\|$ is the unit LOS vector of radar point $\tilde{\boldsymbol{p}}^{R}_j$, and the velocity threshold $\delta_v$ is set to $0.3$\mps.

\subsubsection{Initial Pose Estimation}
The radar inliers $\mathcal{R}_{\mathrm{in}}$ are used to form the radar Doppler residual set in \eqref{eq:RIO} and are optimized jointly with the IMU residuals. The initial pose estimation problem is given as follows:

\small
\begin{equation}
\label{eq:RIO}
\begin{aligned}
    \min_{\boldsymbol{x}} \Biggl\{
    \sum_{k \in \mathcal{I}}
    \Bigl(
    w_\omega \left\|\mathbf{r_{\omega_k}}(t_k)\right\|_2^2
    + &w_a \left\|\mathbf{r_{a_k}}(t_k)\right\|_2^2
    \Bigr)
    \\
    + &\sum_{k \in \mathcal{R}_{\mathrm{in}}}
     w_R \left\|\mathbf{r_{R_k}}(t_k)\right\|_2^2
    \Biggr\}.
\end{aligned}
\end{equation}\normalsize

\begin{figure}[t!]
    \centering
    \includegraphics[width=0.9\columnwidth]{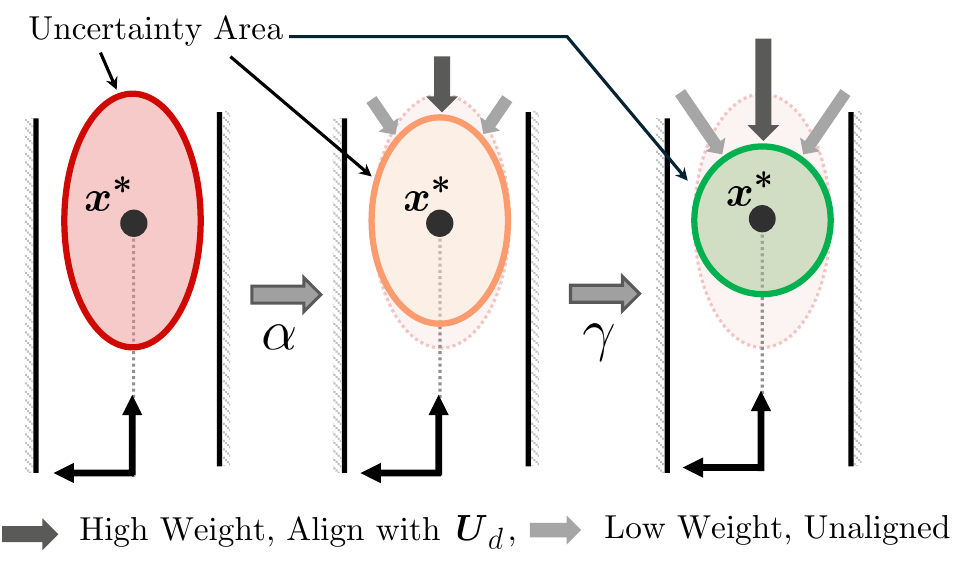}    
  \caption{In tunnel-like scenes, optimization uncertainty becomes anisotropic along weak translational directions (Red); per-point radar reweighting ($\alpha$) and adaptive gain scheduling ($\gamma$) reduce this anisotropy (Yellow, Green).}
  \label{fig:method}
  \vspace{-7mm}
\end{figure}

\subsection{Geometry Only Degeneracy Diagnosis}
\label{sec:degeneracy_diagnosis}
Many \ac{LIO} methods diagnose degeneracy using the eigenvalues or condition number of the Hessian formed from LiDAR point-to-plane residuals~\cite{xicp2024, superloc2025}. In continuous time B-spline \ac{LIO}, however, this can be unreliable because LiDAR residuals with respect to control points are affected not only by scene geometry but also by spline basis coupling and knot placement~\cite{quenzel2025liomars}. As a result, \ac{CP}-space conditioning does not directly reflect geometric degeneracy, and interpreting it as such can lead to incorrect diagnoses.

In this work, degeneracy diagnosis is restricted to translation. 
Orientation is typically well constrained by high rate IMU gyroscope measurements, whereas translation is less directly constrained because accelerometer measurements constrain the second derivative of position rather than the first, and accelerometer errors, including bias and measurement noise, accumulate through double integration, leading to drift in the translational estimate \cite{do2024dero}.  
% Accordingly, instead of relying on the \ac{CP}-space Hessian, we diagnose translational degeneracy using a geometry-only measure computed directly from each LiDAR scan. Following X-ICP~\cite{xicp2024}, we diagnose translation degeneracy using a scan-level $3\times 3$ normal information matrix $\boldsymbol{N}$ computed directly from a single incoming LiDAR scan:
Accordingly, instead of relying on the \ac{CP}-space Hessian, we diagnose translational degeneracy using a scan-level $3\times 3$ normal information matrix $\boldsymbol{N}$, following X-ICP~\cite{xicp2024}:

\small
\begin{equation}
\label{eq:normal-matrix}
\boldsymbol{N} \;\triangleq\; \sum_{i=1}^{M} \boldsymbol{n}_i \boldsymbol{n}_i^\top \in \mathbb{R}^{3\times 3},
\end{equation}\normalsize
where $\boldsymbol{n}_i \in \mathbb{R}^3$ denotes the unit normal vector of the $i$-th planar feature identified in the scan. Planar features are extracted via local neighborhood plane fitting as in FAST-LIO2 \cite{fastlio2}, computed directly within the current scan. Although individual scan-only normals may be noisy, aggregating many such normals provides a stable scan-level degeneracy indicator.

Let the eigendecomposition of $\boldsymbol{N}$ be $\boldsymbol{N} = \boldsymbol{V} \boldsymbol{\Lambda} \boldsymbol{V}^\top$, where $\boldsymbol{\Lambda}=\mathrm{diag}(\lambda_1,\lambda_2,\lambda_3)$ with $\lambda_1\ge \lambda_2\ge \lambda_3\ge 0$, and $\boldsymbol{v}_k$ is the eigenvector of $\lambda_k$. We define the degeneracy subspace in the LiDAR frame by collecting eigen-directions whose eigenvalues are small relative to the dominant one:

\small
\begin{equation}
\label{eq:Ud-definition}
\boldsymbol{U}_d \;\triangleq\; \bigl[ \boldsymbol{v}_k \;\big|\; \lambda_k \le \tau \lambda_1  \bigr] \in \mathbb{R}^{3\times d},
\end{equation}\normalsize
where $\tau\in[0,1]$ is a user-defined ratio threshold and $d\in\{0,1,2\}$ is the resulting subspace dimension, corresponding to weakly informative translational directions. In our experiments, we set $\tau = 0.2$.

\subsection{Degeneracy-Aware Per-Point Radar Reweighting} 
\label{sec:Radar_reweighting} 
Radar Doppler measurements constrain the ego-velocity along each point's \ac{LOS}. When a sufficient number of points are available, these Doppler constraints can complement LiDAR geometric constraints in segments where the LiDAR scan geometry is weak. Since each point contributes a rank-1 directional constraint along its \ac{LOS}, we apply per-point reweighting and assign larger weights to points whose \ac{LOS} directions align with the LiDAR-diagnosed degeneracy subspace, rather than using uniform weights. The effect of per-point reweighting and gain scheduling is illustrated conceptually in \figref{fig:method}.

\subsubsection{Degeneracy Subspace LOS Coverage Check} 
\label{sec:LOS_check} 
As radar can support the LiDAR diagnosed weak subspace only if the current set of radar \ac{LOS} directions provides sufficient coverage of the degeneracy subspace, a lightweight \ac{LOS} coverage check is performed before applying degeneracy-aware reweighting. For each radar measurement $j$, define the unit LOS vector $\tilde{\boldsymbol{l}}_j^{R}\in\mathbb{R}^{3}$ and its projection onto the degeneracy subspace as

\small
\begin{equation}
\begin{aligned}
\label{eq:zj_def}
\boldsymbol{z}_j &= \boldsymbol{U}_d^\top {\boldsymbol{R}}^{{L}}_{R}\; \tilde{\boldsymbol{l}}^{R}_j \in \mathbb{R}^d.
\end{aligned}
\end{equation}\normalsize
Stacking these projected vectors yields the sensitivity matrix

\small
\begin{equation}
\begin{aligned}
\label{eq:Z_def}
\boldsymbol{Z} \;\triangleq\; \left[\boldsymbol{z}_1,\ldots,\boldsymbol{z}_q\right]\in\mathbb{R}^{d\times q},
\end{aligned}
\end{equation}\normalsize
and the corresponding coverage matrix in the degeneracy subspace is defined as

\small
\begin{equation}
\label{eq:M_def}
\boldsymbol{M} \;\triangleq\; \boldsymbol{Z}\boldsymbol{Z}^\top \;=\; \sum_{j=1}^{q}\boldsymbol{z}_j\boldsymbol{z}_j^\top \;\in\;\mathbb{R}^{d\times d}.
\end{equation}\normalsize
By construction, $\boldsymbol{M}$ summarizes how well the current set of radar LOS directions excites the LiDAR-diagnosed degeneracy subspace. In our implementation, observability is checked using a coverage test on $\boldsymbol{M}$: for $d=1$, the subspace is accepted by default, whereas for $d=2$, it is regarded as observable only if $\mathrm{cond}(\boldsymbol{M})<30$. Otherwise, Doppler constraints are not considered sufficient to support all weak axes; therefore, degeneracy-aware reweighting is disabled for the current scan. To improve \ac{LOS} coverage of the LiDAR diagnosed degeneracy subspace, the radar was mounted with a 20$^\circ$ downward tilt, similar to prior radar setups~\citep{kulkarni2025unipilot}.

\subsubsection{Calculate Per-Point Weight}
If $\boldsymbol{M}$ sufficiently excites the degeneracy subspace, a non-negative weight $\alpha_j$ is computed for each radar point $j$. Denoting the radar frame ego-velocity implied by the state $\boldsymbol{x}$ at time $t_r$ as $v^{R}(\cdot)$, the radar Doppler residual for point $j$ is expressed as
\begin{equation}
{\begin{aligned}
\label{eq:doppler-residual}
\mathbf{r_R}_j(t_r) &= -\tilde{\boldsymbol{l}}^{R^\top}_j\, \boldsymbol{y}(\boldsymbol{x}) - \tilde{v}^{R}_{m_j},\qquad
\boldsymbol{y} \triangleq \boldsymbol{v}^{R}(t_r;\boldsymbol{x})\in\mathbb{R}^3.
\end{aligned}}
\end{equation}
Applying first-order linearization to \eqref{eq:doppler-residual} yields

\small
\begin{equation}
\label{eq:doppler-linearization}
\delta \mathbf{r_R}_j \approx -\tilde{\boldsymbol{l}}^{R^\top}_j\delta \boldsymbol{y}.
\end{equation}\normalsize
Stacking all Doppler residuals results in $\delta \boldsymbol{r_R} \approx -\boldsymbol{L}\delta \boldsymbol{y}$, where $\boldsymbol{L} \triangleq [\tilde{\boldsymbol{l}}^{R}_1,\ldots,\tilde{\boldsymbol{l}}^{R}_q]^\top \in \mathbb{R}^{q\times 3}$. With the diagonal weight matrix $\boldsymbol{W}_R(\alpha)\triangleq \mathrm{diag}(w_{R}\alpha_1,\ldots,w_{R}\alpha_q)$, the induced weighted least squares term in the cost is

\small
\begin{equation}
\label{eq:doppler-quad-term}
\begin{aligned}
\left\| \delta \boldsymbol{r}_R \right\|_{\boldsymbol{W}_R}^{2}
\triangleq\delta \mathbf{r_R}^\top \boldsymbol{W}_R(\alpha) \delta \mathbf{r_R} 
&= \delta \boldsymbol{y}^\top \underbrace{(\boldsymbol{L}^\top \boldsymbol{W}_R(\alpha) \boldsymbol{L})}_{\boldsymbol{D}(\alpha)} \delta \boldsymbol{y} \\
\boldsymbol{D}(\alpha) &= \sum_{j=1}^{q} \bigl(w_{R}\, \alpha_j\bigr) \; \tilde{\boldsymbol{l}}^{R}_j \; \tilde{\boldsymbol{l}}^{R^\top}_j.
\end{aligned}
\end{equation}\normalsize
Each point contributes a rank-1 term $\tilde{\boldsymbol{l}}^{R}_j \tilde{\boldsymbol{l}}^{R^\top}_j$ to the Doppler information matrix $\boldsymbol{D}(\alpha)$. We set $\alpha_j$ based on the alignment between the point \ac{LOS} and the LiDAR diagnosed weak subspace to increase Doppler information in $\boldsymbol{U}_d$. With the projection vector $\boldsymbol{z}_j$ in \eqref{eq:zj_def}, the alignment score $a_j$ is defined as:

\small
\begin{equation}
\begin{aligned}
\label{eq:alignment-score}
a_j &\triangleq\; \|\boldsymbol{z}_j\|^2 \;=\; \boldsymbol{z}_j^\top \boldsymbol{z}_j,
\end{aligned}
\end{equation}\normalsize
and the per-point weights are given by an exponential function normalized to unit mean:

\small
\begin{equation}
\label{eq:exp-weights}
\alpha_j \;=\;\frac{\exp(\eta\, a_j)}{\frac{1}{q}\sum_{k=1}^{q}\exp(\eta\, a_k)},
\qquad \eta>0,
\end{equation}\normalsize
where $q$ denotes the number of radar points and $\eta$ controls the distribution sharpness. This reweighting emphasizes points aligned with the degeneracy subspace while reducing the influence of points along directions already constrained by the LiDAR scan. The value of $\eta$ is selected empirically, and its effect is evaluated in the ablation study.
\input{tab/table1.tex}
\subsection{Radar Gain Scheduling}
\label{sec:Radar_gain_scheduling}
The per-point weights $\alpha_j$ shape the directional distribution of the Doppler information but not its overall scale. Because the relative balance between weak and well-constrained translational directions varies with the LiDAR scan geometry, we introduce a global gain $\gamma$ to adjust the radar Doppler contribution per scan. To schedule $\gamma$, we use the sphericity measure $S(\boldsymbol{N})$ of the LiDAR plane normal matrix $\boldsymbol{N}$, computed per scan to summarize the imbalance of geometric constraints across translational directions:

\small
\begin{equation}
\begin{aligned}
\label{eq:gamma-schedule}
S(\boldsymbol{N}) \;\triangleq\;&\frac{\bigl(\det \boldsymbol{N}\bigr)^{1/3}}{\tfrac{1}{3}\operatorname{tr}(\boldsymbol{N})}
\;=\;
\frac{\left(\lambda_1\lambda_2\lambda_3\right)^{1/3}}{\tfrac{1}{3}\left(\lambda_1+\lambda_2+\lambda_3\right)} \\
\gamma =\;& g\!\left(S(\boldsymbol{N})\right)=1/\max\!\left(S(\boldsymbol{N}),10^{-3}\right)
\end{aligned}
\end{equation}\normalsize
where $\lambda_1 \ge \lambda_2 \ge \lambda_3$ are the eigenvalues of $\boldsymbol{N}$. The sphericity $S(\boldsymbol{N})$ approaches one for isotropic geometry and decreases as the information becomes concentrated in fewer directions. We then use a monotone mapping $g(\cdot)$ so that $\gamma$ increases as $S(\boldsymbol{N})$ decreases, giving more weight to radar Doppler under anisotropic LiDAR geometry.

After computing the Doppler weights $\{\alpha_j\}$ and gain $\gamma$, state estimation is carried out by solving the tightly coupled nonlinear least squares problem in \eqref{eq:tightly-coupled}. In the radar Doppler term $\mathbf{r_{R_k}}(\cdot)$, when degeneracy-aware reweighting is enabled (\ie, the translation degeneracy check in \Cref{sec:degeneracy_diagnosis} and \ac{LOS} coverage check in \Cref{sec:LOS_check} pass), 
% we replace the nominal weight $w_R$ with an effective weight
the effective radar Doppler weight $w_{R,k}^{\textbf{*}}$ in \eqref{eq:tightly-coupled} is defined as

\small
\begin{equation}
\label{eq:wR_eff}
w_{R,k}^{\textbf{*}} \triangleq w_R\,\gamma\,\alpha_k,
\end{equation}\normalsize
and otherwise set $w_{R,k}^{\textbf{*}} \triangleq w_R$. 
% Therefore, the radar Doppler term in \eqref{eq:tightly-coupled} can be written as $\sum_{k \in \mathcal{R}} w_{R,k}^{\textbf{*}}\,\left\|\mathbf{r_{R_k}}(t_k)\right\|_2^2.$

% \begin{figure}[!t]
%   \centering
%   \includegraphics[width=\columnwidth]{figure/RLIOvsLIO(down).pdf}
%   \caption{\textbf{Performance change from adding Radar.} For each sequence, the ATE and RTE ratios (RLIO/LIO) are reported; values $<1$ indicate improvement and values $>1$ indicate degradation.}
%   \label{fig:Radar_effect}
%   \vspace{-7mm}
% \end{figure}

%% file: tab/table1.tex
\definecolor{cellgreenA}{RGB}{230,238,195}
\definecolor{cellgreenB}{RGB}{199,228,207}
\newcommand{\err}[2]{\makebox[2.5em][r]{#1}\,/\,\makebox[2.7em][l]{#2}}
\begin{table*}[t]
\centering
\caption{ATE (RMSE) [\meter] / RTE (RMSE) [\meter/\meter]. (\colorbox{cellgreenB}{\textbf{Bold}}: Best, \colorbox{cellgreenA}{\underline{Underline}}: Second-Best, $\times$ : fail)}
\label{tab:rmse_eval}
\setlength{\tabcolsep}{4pt}
\renewcommand{\arraystretch}{1.45}
\resizebox{\textwidth}{!}{
\begin{tabular}{c|c|*{7}{c}|*{6}{c}}
\toprule
\multicolumn{2}{c|}{}&
\multicolumn{7}{c|}{\textbf{GaRLILEO Dataset}} &
\multicolumn{6}{c}{\textbf{In-house Dataset}} \\
\cmidrule(r){1-2}\cmidrule(lr){3-9}\cmidrule(l){10-15}

\multirow{2}{*}{\textbf{Method}}
& \multirow{2}{*}{\textbf{Type}}
& \textbf{Downstair} & \textbf{CorriLoop} & \textbf{BiCorridor}
& \textbf{SlopeStair} & \textbf{Tunnel} & \textbf{Overpass} & \textbf{Quad}
& \textbf{BikeTunnel1} & \textbf{BikeTunnel2} & \textbf{Park1}
& \textbf{Park2} & \textbf{Airfield1} & \textbf{Airfield2} \\

& &
\unit{233.75}{\meter} & \unit{208.68}{\meter} & \unit{240.82}{\meter} & \unit{273.37}{\meter} & \unit{247.94}{\meter} & \unit{169.17}{\meter} & \unit{447.83}{\meter}
& \unit{360.34}{\meter} & \unit{153.84}{\meter} & \unit{159.60}{\meter}
& \unit{265.17}{\meter} & \unit{198.98}{\meter} & \unit{208.65}{\meter} \\
\midrule\midrule

\texttt{GenZ-ICP}
& LO
& \err{1.208}{0.055} & \err{0.728}{0.063} & \err{0.569}{0.057} & \err{1.966}{0.050} & \err{0.891}{0.040} & \err{1.346}{0.062} & \err{4.239}{0.042}
& $\times$ & $\times$ & $\times$ & $\times$ & $\times$ & $\times$ \\

\texttt{FAST-LIO2}
& LIO
& \err{0.459}{\colorbox{cellgreenA}{0.036}} & \err{0.336}{0.049} & \err{0.767}{0.040} & \err{\colorbox{cellgreenA}{0.867}}{\colorbox{cellgreenA}{0.030}} & \err{\colorbox{cellgreenA}{0.525}}{0.031} & \err{1.067}{0.036} & \err{3.616}{0.040}
& $\times$ & \err{66.072}{1.829} & \err{2.296}{0.234} & \err{11.354}{0.299} & \err{0.681}{0.216} & \err{1.894}{0.194} \\

\texttt{SuperOdom}
& LIO
& \err{1.011}{0.057} & \err{2.927}{0.219} & \err{4.372}{0.376} & \err{1.531}{0.043} & \err{3.579}{0.132} & \err{1.286}{0.061} & \err{3.665}{0.044}
& $\times$ & $\times$ & $\times$ & $\times$ & $\times$ & $\times$ \\

\texttt{DLIO}
& LIO
& \err{0.887}{0.059} & \err{0.284}{0.064} & \multicolumn{1}{c}{$\times$}
& \err{\colorbox{cellgreenB}{\textbf{0.694}}}{0.065} & \err{0.791}{0.055} & \err{0.954}{0.067} & \err{3.464}{0.062}
& $\times$ & $\times$ & \err{1.008}{0.162} & \err{\colorbox{cellgreenB}{\textbf{1.219}}}{0.159} & \err{0.627}{0.141} & \err{0.942}{0.139} \\

\texttt{RESPLE}
& LIO
& \err{0.471}{0.044} & \err{0.152}{0.050} & \err{0.508}{0.073} & \err{1.183}{0.042} & \err{1.214}{0.030} & \err{0.912}{0.040} & $\times$
& $\times$ & $\times$ & $\times$ & $\times$ & \err{\colorbox{cellgreenA}{0.611}}{0.115} & \err{0.464}{0.110} \\

\texttt{DR-LRIO}
& RLIO
& \err{0.565}{0.045} & \err{0.444}{0.053} & \err{1.051}{\colorbox{cellgreenA}{0.039}} & \err{2.242}{0.044} & \err{1.700}{0.037}
& \err{\colorbox{cellgreenB}{\textbf{0.621}}}{0.046} & \err{\colorbox{cellgreenA}{2.786}}{0.044}
& \err{\colorbox{cellgreenA}{1.881}}{\colorbox{cellgreenA}{0.069}} & \err{\colorbox{cellgreenA}{0.605}}{\colorbox{cellgreenA}{0.097}} & \err{\colorbox{cellgreenA}{0.853}}{0.073} & \err{1.820}{0.061} & \err{1.399}{0.062} & \err{2.485}{0.067} \\

% \texttt{Mimosa}
% & LIO
% & \err{0.578}{0.037} & \err{\colorbox{cellgreenB}{\textbf{0.190}}}{0.054} & \err{\colorbox{cellgreenB}{\textbf{0.299}}}{0.040}
% & \err{\colorbox{cellgreenB}{\textbf{0.985}}}{\colorbox{cellgreenB}{\textbf{0.035}}} & \err{\colorbox{cellgreenB}{\textbf{0.764}}}{\colorbox{cellgreenB}{\textbf{0.034}}} & \err{0.813}{\colorbox{cellgreenB}{\textbf{0.038}}} & \err{3.516}{\colorbox{cellgreenB}{\textbf{0.043}}}
% & \err{0.000}{0.000} & \err{0.000}{0.000} & \err{0.000}{0.000} & \err{0.000}{0.000} & \err{0.000}{0.000} & \err{0.000}{0.000} \\

\texttt{GaRLIO}
& RLIO
& \err{0.225}{0.147} & \err{0.223}{0.055} & \multicolumn{1}{c}{$\times$}
& \err{5.040}{0.745} & \err{\colorbox{cellgreenB}{\textbf{0.506}}}{0.117} & \multicolumn{1}{c}{$\times$} & $\times$
& \err{2.507}{0.097} & \err{1.558}{0.250} & \err{\colorbox{cellgreenB}{\textbf{0.423}}}{\colorbox{cellgreenA}{0.035}} & \err{4.602}{\colorbox{cellgreenB}{\textbf{0.039}}} & \err{0.773}{\colorbox{cellgreenA}{0.028}} & \err{0.580}{\colorbox{cellgreenA}{0.025}} \\

\midrule

\texttt{\bfseries Ours-LIO}
& LIO
& \err{\colorbox{cellgreenB}{\textbf{0.202}}}{0.037} & \err{\colorbox{cellgreenB}{\textbf{0.128}}}{\colorbox{cellgreenA}{0.045}} & \err{\colorbox{cellgreenA}{0.322}}{0.041}
& \err{1.423}{0.031} & \err{1.085}{\colorbox{cellgreenA}{0.028}} & \err{0.836}{\colorbox{cellgreenB}{\textbf{0.032}}} & \err{2.946}{{\colorbox{cellgreenB}{\textbf{0.036}}}}
& $\times$ & \err{12.352}{0.546} & $\times$ & $\times$ &  \err{9.627}{0.180} &  \err{\colorbox{cellgreenB}{\textbf{0.127}}}{0.027}\\

\texttt{\bfseries Ours-RLIO}
& RLIO
& \err{\colorbox{cellgreenA}{0.213}}{\colorbox{cellgreenB}{\textbf{0.035}}} & \err{\colorbox{cellgreenA}{0.139}}{\colorbox{cellgreenB}{\textbf{0.044}}}
& \err{\colorbox{cellgreenB}{\textbf{0.279}}}{\colorbox{cellgreenB}{\textbf{0.036}}}
& \err{1.374}{\colorbox{cellgreenB}{\textbf{0.029}}} & \err{1.017}{\colorbox{cellgreenB}{\textbf{0.028}}} & \err{\colorbox{cellgreenA}{0.813}}{\colorbox{cellgreenA}{0.034}} & \err{\colorbox{cellgreenB}{\textbf{2.374}}}{\colorbox{cellgreenA}{0.036}}
& \err{\colorbox{cellgreenB}{\textbf{1.532}}}{\colorbox{cellgreenB}{\textbf{0.050}}} & \err{\colorbox{cellgreenB}{\textbf{{0.485}}}}{\colorbox{cellgreenB}{\textbf{0.047}}} & \err{1.010}{\colorbox{cellgreenB}{\textbf{0.033}}} & \err{\colorbox{cellgreenA}{1.524}}{\colorbox{cellgreenA}{0.047}} & \err{\colorbox{cellgreenB}{\textbf{0.183}}}{\colorbox{cellgreenB}{\textbf{0.026}}} & \err{\colorbox{cellgreenA}{0.142}}{\colorbox{cellgreenB}{\textbf{0.024}}} \\

\bottomrule
\end{tabular}}
\vspace{-5mm}
\end{table*}

%% file: sections/4_experiment.tex
\section{experiment}
\subsection{Datasets and Evaluation Metric}
\label{sec:experiment}

We evaluate \textbf{TRaIL-Odom} against several \ac{SOTA} state estimation algorithms, including LiDAR-based methods: \texttt{GenZ-ICP}~\cite{lee2024genz}, \texttt{FAST-LIO2}~\cite{fastlio2}, \texttt{SuperOdom}~\cite{zhao2021superodom}, \texttt{DLIO}~\cite{kenny2023dlio}, and \texttt{RESPLE}~\cite{cao2025resple}; and radar-LiDAR fusion methods: \texttt{DR-LRIO}~\cite{nissov2024Degrad} and \texttt{GaRLIO}~\cite{cynoh2025garlio}. \texttt{Ours-LIO} denotes the LiDAR-IMU-only variant of our framework. It retains the continuous-time B-spline trajectory representation and graph-based tightly coupled sliding-window optimization of LiDAR point-to-plane and IMU residuals, while disabling all radar-related components.

\smallskip\noindent\textbf{Datasets.} We use two datasets with different geometric characteristics. The first is the GaRLILEO dataset~\cite{noh2025garlileo}, which contains well-constrained and semi-degenerate sequences in constructed indoor and outdoor environments. The second is our in-house dataset, collected with Boxi~\cite{frey2025boxi} to evaluate performance under optimization degeneracy; it consists of two sequences of different lengths in each of three environments. 

\begin{figure}[!t]
  \centering
  \includegraphics[width=0.9\columnwidth]{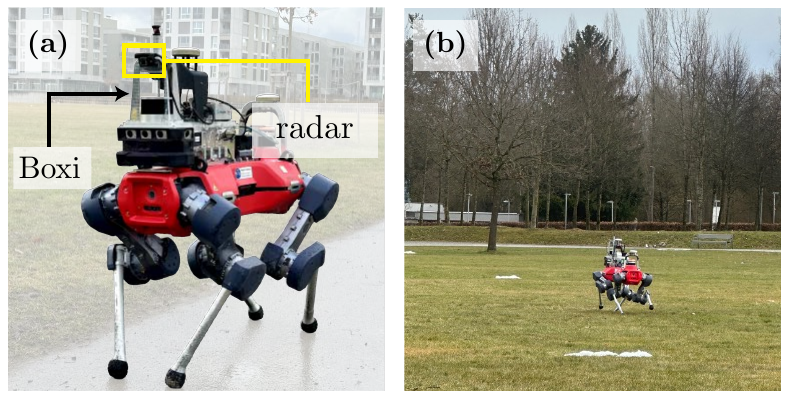}
  \caption{(a) Sensing platform; (b) In-house dataset scene (\textit{Park*}).}
  \label{fig:Dataset_detail}
  \vspace{-5mm}
\end{figure}

For the in-house dataset, the Boxi platform and environments are shown in \figref{fig:Dataset_detail} and \figref{fig:Teaser}; all sequences were collected using the ANYmal robot. The dataset includes one radar, two LiDARs, ten RGB cameras, and seven IMUs. Ground truth was obtained with a Leica MS60 total station and GRZ101 360$^\circ$ Mini Prism. Intervals with unreliable prism tracking due to \ac{LOS} loss were excluded from evaluation. Details are available on the dataset website\footnote{\texttt{\href{https://chiyunnoh.github.io/TRaIL-Odom/}{https://chiyunnoh.github.io/TRaIL-Odom/}}}. All methods use the Livox Mid-360 LiDAR, Honeywell HG4930 IMU, and D3 Embedded RS-1843AOPU mmWave radar.

\smallskip\noindent\textbf{Evaluation.}
Across both datasets, results are evaluated with the \textit{evo} library~\cite{grupp2017evo} using the \ac{ATE} and \ac{RTE} (over 1-meter segments). Evaluation results are summarized in \tabref{tab:rmse_eval}.

\subsection{In-house Dataset}
As shown in \tabref{tab:rmse_eval} and \figref{fig:Teaser}, the LO baseline fails on all six sequences, and even \ac{SOTA} LIO methods either diverge or exhibit large drift. An exception is \textit{Airfield*}, where a vertical antenna and distant parked vehicles provide sparse but usable features, allowing several LiDAR-based methods to succeed. One notable observation is that \texttt{DLIO} does not exhibit large drift on four of the six in-house sequences and even attains the lowest ATE on \textit{Park2}. This is consistent with its observer-based design~\cite{kenny2023dlio}, which can prevent global drift, but under weak geometry the trajectory still shows local jitter, yielding an RTE about 3.4$\times$ larger than that of \textbf{Ours-RLIO}. This highlights the need for additional radar constraints to improve local consistency in featureless environments.

Radar-aided methods are the only approaches that succeed on all sequences and deliver consistently low errors. \textbf{Ours-RLIO} achieves the best RTE on five out of six sequences, and also attains the lowest ATE on \textit{BikeTunnel1/2} and \textit{Airfield1}. The only exception is \textit{Park2}, where \texttt{GaRLIO} achieves the lowest RTE but with a substantially larger ATE, whereas \textbf{Ours-RLIO} maintains both low RTE and a lower ATE.

\subsection{GaRLILEO Dataset}
On the GaRLILEO dataset, we evaluate seven sequences. Since the dataset is semi-degenerate, most methods succeed on most sequences using sparse geometric features. However, sequences such as \textit{BiCorridor} and \textit{Quad} start with narrow corridors and stair, which increase geometric degeneracy and cause several LiDAR-based methods to fail. In addition, \texttt{GaRLIO} fails when its two-stage updates are affected by a high outlier ratio in sparse radar point clouds during sweep preprocessing. Notably, Ours-LIO shows competitive performance compared with existing LIO baselines. This is mainly due to its graph-based tightly coupled sliding-window LiDAR-IMU optimization with a continuous-time B-spline trajectory representation. The optimizer jointly refines active spline control points and IMU biases, while the marginalization prior preserves information from the previous window.

Building on this \ac{LIO} baseline, adding radar yields comparable or improved accuracy, as shown in \tabref{tab:radar_ratio}. In particular, the radar-aided mode improves \ac{ATE} on five out of seven sequences, with clear gains on \textit{BiCorridor} and \textit{Quad}. This suggests that direction-dependent radar weighting reinforces weakly observable directions without unnecessarily affecting well-constrained ones. The same trend appears in \ac{RTE}: ours improves local accuracy on four sequences, is unchanged on two, and degrades only slightly on \textit{Overpass}. In contrast, \texttt{DR-LRIO} does not adaptively reallocate radar contribution, and adding radar degrades \ac{ATE} by factors of 2.2-3.5 on \textit{CorriLoop}, \textit{BiCorridor}, \textit{SlopeStair}, and \textit{Tunnel}. It also degrades \ac{RTE} on five out of seven sequences, resulting in a worse mean ratio of 1.102.

\input{tab/table2.tex}

\begin{figure}[!t]
  \centering
  \includegraphics[width=0.95\columnwidth]{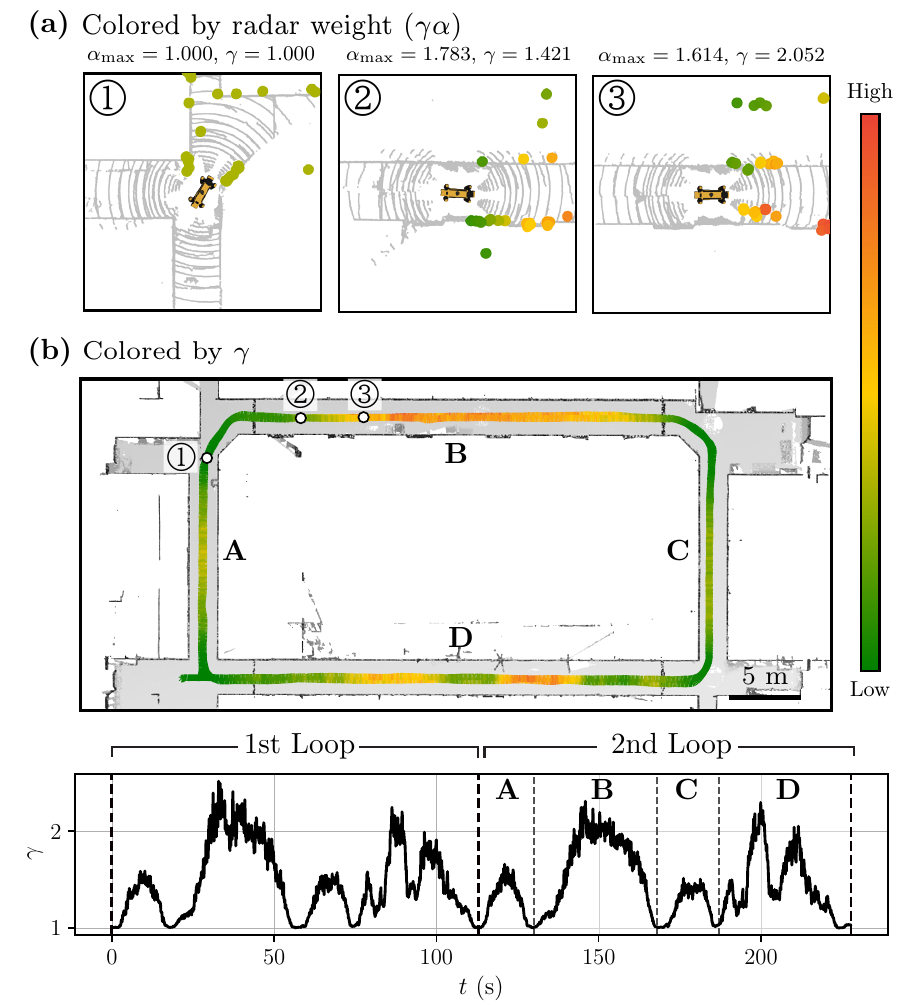}
  \caption{\textbf{Adaptive radar weighting on the \textit{CorriLoop} sequence.} (a) Radar measurements colored by the effective weight ($\alpha\; \&\; \gamma$) at three representative locations. (b) Trajectory colored by the scan-wise gain ($\gamma$) on the ground-truth map, along with its temporal evolution. A–D denote corridor segments, and \ding{192}-\ding{194} mark the locations of the examples in (a).}
  \label{fig:CorriLoop_plot}
  \vspace{-7mm}
\end{figure}

\subsection{Degeneracy-Aware Behavior of Adaptive Radar Weighting}
To better understand the proposed adaptive weighting, we analyze a representative \textit{CorriLoop} sequence with repeated transitions between geometrically informative and degenerate regions. As shown in \figref{fig:CorriLoop_plot}, the scan-wise gain $\gamma$ increases in corridor segments where LiDAR geometry is weak, while remaining close to unity in more informative regions such as corners. Meanwhile, the per-point factor $\alpha$ exhibits a non-uniform spatial distribution, assigning larger effective weights to radar measurements that better support weakly observable motion directions. 

Moreover, the temporal profile of $\gamma$ remains similar across repeated traversals of the same corridor segments. In particular, the first and second loops show consistent increases and decreases of $\gamma$ over corresponding regions, suggesting that the adaptation is primarily driven by scene geometry. This repeatability supports the interpretation that $\gamma$ captures environment-dependent degeneracy in a stable manner.

\subsection{Ablation Study}
\subsubsection{Effect on each module}
Table~\ref{tab:system_ablation} shows that the two weighting modules (\Cref{sec:Radar_reweighting} and \Cref{sec:Radar_gain_scheduling}) play different roles. With only radar gain scheduling enabled ($\gamma$), both \ac{ATE} and \ac{RTE} decrease relative to the fixed-weight baseline on all sequences, indicating that increasing the overall radar Doppler contribution under anisotropic LiDAR geometry is effective.
With only per-point reweighting enabled ($\alpha$), \ac{ATE} and \ac{RTE} also improve, but more selectively across sequences. This is consistent with the role of $\alpha$ as a directional redistribution mechanism: it reallocates Doppler information across translational directions but does not increase its overall strength. 
\input{tab/table3.tex}
\input{tab/table4.tex}

\begin{figure}[!t]
  \centering
  \includegraphics[width=\columnwidth]{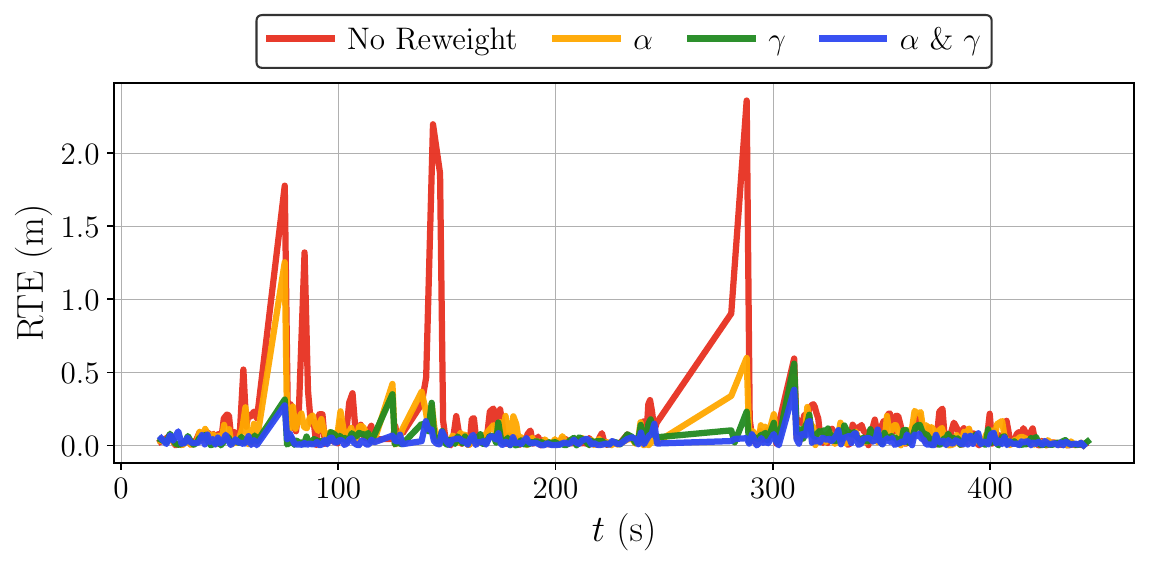}
  \caption{\textbf{RTE over time on \textit{BikeTunnel1}.} RTE is plotted for four variants: no reweighting, $\alpha$ only (per-point reweighting), $\gamma$ only (adaptive gain), and $\alpha\; \&\; \gamma$ (full method).}
  \label{fig:rte_over_time}
  \vspace{-7mm}
\end{figure}

\begin{figure}[!t]
  \centering
  \includegraphics[width=\columnwidth]{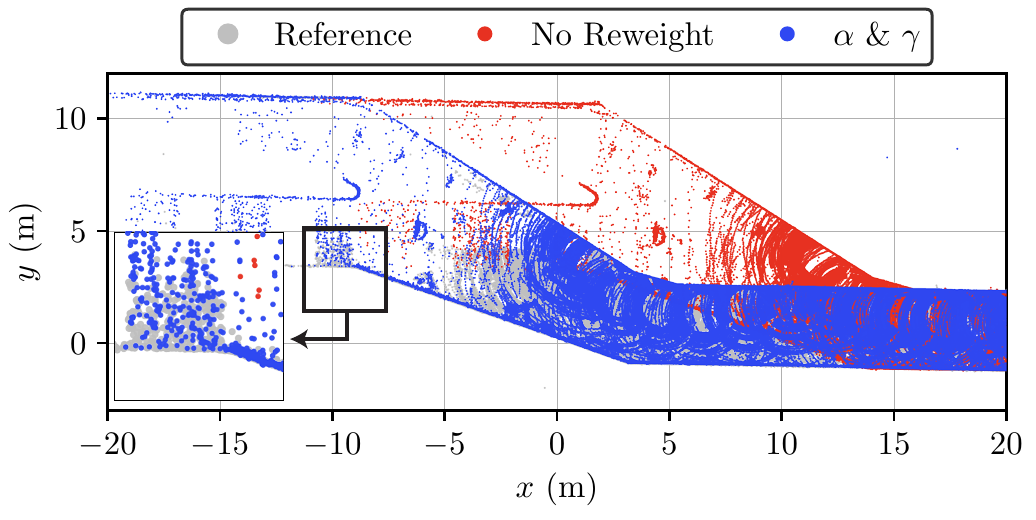}
  \caption{\textbf{Return-to-start translation drift on \textit{BikeTunnel2}.} Comparison between fixed-weight fusion (Red) and the full method (Blue, $\alpha\; \&\; \gamma$). The full method reduces the return-to-start drift from \unit{10.48}{\m} to \unit{0.14}{\m}}
  \label{fig:qualitative image}
  \vspace{-7mm}
\end{figure}

When both modules are enabled, \ac{RTE} is further reduced on all sequences compared to using either module alone, indicating that $\alpha$ allocates the additional curvature introduced by $\gamma$ toward weak directions more effectively, which yields larger improvements. For \ac{ATE}, the combined setting improves \textit{BikeTunnel1} and \textit{Airfield1}, while \textit{Park1} shows a lower ATE with $\gamma$ only. Overall, $\gamma$ controls when radar Doppler should contribute more, and $\alpha$ controls where that contribution is directed. This is also visible in \figref{fig:rte_over_time} and \figref{fig:qualitative image}. In \figref{fig:rte_over_time}, the fixed-weight baseline shows four pronounced RTE peaks caused by geometric degeneracy. These peaks are reduced with $\alpha$, indicating that directional redistribution helps but is insufficient on its own. Using only $\gamma$ suppresses peaks more effectively by increasing the overall radar information strength under weak LiDAR geometry, while the full method ($\alpha+\gamma$) yields the most stable estimate throughout the sequence. \figref{fig:qualitative image} further illustrates this effect through the return-to-start drift. The fixed-weight result shows a drift of 10.48\,m, whereas the full method reduces it to only 0.14\,m, demonstrating a substantial improvement in performance.

\input{tab/table5.tex}

\subsubsection{Effect on \texorpdfstring{$\eta$}{eta} parameter}
Table~\ref{tab:eta_tunnel5_plain} evaluates the sensitivity of the per-point radar reweighting in \eqref{eq:exp-weights} to the sharpness parameter $\eta$ on \textit{BikeTunnel2}. Recall that $\eta$ controls how strongly the weights $\{\alpha_j\}$ concentrate on radar points whose \ac{LOS} directions align better with the LiDAR diagnosed weak subspace. When $\eta=0$, reweighting is disabled, and all points contribute uniformly. As $\eta$ increases from 0, ATE, RTE, and RTE$_{\max}$ generally decrease, indicating that allocating radar Doppler information toward the weak directions improves both global and local accuracy. However, when $\eta$ becomes too large, performance degrades. This is consistent with weights becoming overly concentrated on a small subset of points, thereby increasing sensitivity to remaining outliers.
Given this insight, the lowest \ac{ATE} is obtained at $\eta=3$, while \ac{RTE} and \ac{RTE}$_{\max}$ attain their minimum at $\eta=2$. Based on the relative error metrics, we use $\eta=2$ in our algorithm.

\subsection{Runtime Analysis}
Table~\ref{tab:runtime_modules} reports the average per-frame runtime of each component, including its required preprocessing time, such as LiDAR feature extraction and scan-to-map association in ``Optim.'' and normal estimation in ``Reweight''.
Experiments were conducted on an Intel Core i9-14900HX CPU with 64 GB of RAM. The joint optimization dominates the computational cost, requiring about \unit{40-51}{\ms} per LiDAR scan, while the proposed degeneracy-aware reweighting modules introduce only a small additional overhead of \unit{0.93-2.95}{\ms}. Across all evaluated sequences, the total runtime remains below \unit{100}{\ms} per scan, satisfying the real-time requirement for our \unit{10}{\hertz} LiDAR setup.

%% file: tab/table2.tex
\definecolor{meanbg}{gray}{0.93}
\definecolor{cellgreenB}{RGB}{199,228,207}

\newcommand{\ratiofmt}[1]{%
  \begingroup
  \pgfmathtruncatemacro{\ratioflag}{(#1 <= 1.0005) ? 1 : 2}%
  \ifnum\ratioflag=1
    \cellcolor{cellgreenB}\textbf{#1}%
  \else
    #1%
  \fi
  \endgroup
}

\begin{table}[t]
\centering
\caption{\textbf{Performance change from adding radar.} For each sequence, the ATE and RTE ratios (RLIO/LIO) are reported; values $<1$ indicate improvement and values $>1$ indicate degradation. (\colorbox{cellgreenB}{\textbf{Bold}}: Improvement)}
\label{tab:radar_ratio}
\setlength{\tabcolsep}{4.5pt}
\scriptsize
\begin{tabular}{c|cc!{\color{black!35}\vrule width 0.3pt}cc!{\color{black!35}\vrule width 0.3pt}cc!{\color{black!35}\vrule width 0.3pt}cc}
\toprule
\multirow{2}{*}{Method}
& \multicolumn{2}{c!{\color{black!35}\vrule width 0.3pt}}{\textbf{Downstair}}
& \multicolumn{2}{c!{\color{black!35}\vrule width 0.3pt}}{\textbf{CorriLoop}}
& \multicolumn{2}{c!{\color{black!35}\vrule width 0.3pt}}{\textbf{BiCorridor}}
& \multicolumn{2}{c!{\color{black!35}\vrule width 0.3pt}}{\textbf{SlopeStair}} \\
& ATE & RTE & ATE & RTE & ATE & RTE & ATE & RTE \\
\midrule
\texttt{\bfseries DR-LRIO}
& \ratiofmt{0.978} & \ratiofmt{1.216}
& \ratiofmt{2.337} & \ratiofmt{0.981}
& \ratiofmt{3.515} & \ratiofmt{0.975}
& \ratiofmt{2.276} & \ratiofmt{1.257} \\
\texttt{\bfseries Ours}
& \ratiofmt{1.054} & \ratiofmt{0.946}
& \ratiofmt{1.086} & \ratiofmt{0.978}
& \ratiofmt{0.866} & \ratiofmt{0.878}
& \ratiofmt{0.966} & \ratiofmt{0.935} \\
\midrule
\multirow{2}{*}{Method}
& \multicolumn{2}{c!{\color{black!35}\vrule width 0.3pt}}{\textbf{Tunnel}}
& \multicolumn{2}{c!{\color{black!35}\vrule width 0.3pt}}{\textbf{Overpass}}
& \multicolumn{2}{c!{\color{black!35}\vrule width 0.3pt}}{\textbf{Quad}}
& \multicolumn{2}{c}{\textbf{Mean}} \\
& ATE & RTE & ATE & RTE & ATE & RTE
& ATE & RTE \\
\midrule
\texttt{\bfseries DR-LRIO}
& \ratiofmt{2.225} & \ratiofmt{1.088}
& \ratiofmt{0.764} & \ratiofmt{1.211}
& \ratiofmt{0.792} & \ratiofmt{1.023}
& \ratiofmt{1.580} & \ratiofmt{1.102} \\
\texttt{\bfseries Ours}
& \ratiofmt{0.937} & \ratiofmt{1.000}
& \ratiofmt{0.972} & \ratiofmt{1.062}
& \ratiofmt{0.806} & \ratiofmt{1.000}
& \ratiofmt{0.951} & \ratiofmt{0.970} \\
\bottomrule
\end{tabular}
\vspace{-5mm}
\end{table}

%% file: tab/table3.tex
\begin{table}[t]
\centering
\caption{Ablation study on system components. \textbf{Bold} is the best.}
\label{tab:system_ablation}
\resizebox{0.9\columnwidth}{!}{%
\begin{tabular}{cc|cc|cc|cc}
\toprule
\multicolumn{2}{c|}{Weighting} &
\multicolumn{2}{c|}{\textbf{BikeTunnel1}} &
\multicolumn{2}{c|}{\textbf{Park1}} &
\multicolumn{2}{c}{\textbf{Airfield1}} \\
$\alpha$ & $\gamma$ &
ATE & RTE & ATE & RTE & ATE & RTE \\ 
\midrule
\xmark & \xmark & 6.957 & 0.288 & 7.792 & 0.144 & 4.670 & 0.075 \\
\cmark & \xmark & 3.959 & 0.128 & 2.993 & 0.079 & 3.478 & 0.073 \\
\xmark & \cmark & 2.686 & 0.074 & \textbf{0.802} & 0.036 & 0.231 & 0.034 \\
\cmark & \cmark & \textbf{1.532} & \textbf{0.050} & 1.010 & \textbf{0.033} & \textbf{0.183} & \textbf{0.026} \\
\bottomrule
\end{tabular}%
}
\vspace{-1mm}
\end{table}

%% file: tab/table4.tex
\begin{table}[t]
\centering
\caption{Effect of $\eta$ on \textit{BikeTunnel2}. \textbf{Bold} is the best.}
\label{tab:eta_tunnel5_plain}
\setlength{\tabcolsep}{4pt}
\renewcommand{\arraystretch}{1.15}

\begin{tabular}{c|ccccc cc}
\toprule
$\eta$ & 0.0 & 0.5 & 1.25 & 1.75 & 2 & 3 & 4 \\
\midrule
ATE          & 4.214 & 1.081 & 0.868 & 0.638 & 0.485 & \textbf{0.386} & 0.650 \\
RTE          & 0.210 & 0.073 & 0.061 & 0.051 & \textbf{0.047} & 0.050 & 0.062 \\
RTE$_{\max}$ & 0.563 & 0.222 & 0.205 & 0.163 & \textbf{0.154} & 0.227 & 0.405 \\
\bottomrule
\end{tabular}
\vspace{-1mm}
\end{table}

%% file: tab/table5.tex
\begin{table}[!t]
\centering
\setlength{\tabcolsep}{5pt}
\caption{
{\textbf{Average per-frame runtime of each pipeline component.}
All values are reported in milliseconds. Reweight refers to \secref{sec:degeneracy_diagnosis}, \secref{sec:Radar_reweighting}, and \secref{sec:Radar_gain_scheduling}; Optim. refers to \secref{sec:joint_optimization}.}
}
\label{tab:runtime_modules}
{\begin{tabular}{l|ccccc}
\toprule
Sequence & RI-Init & Reweight & Optim. & Mapping & Total \\
\midrule
\textbf{BikeTunnel1} & 4.64 & 0.93 & 40.72 & 0.39 & 46.68 \\
\textbf{Park1} & 4.46 & 1.65 & 51.34 & 0.84 & 58.29 \\
\textbf{BiCorridor} & 5.04 & 1.67 & 44.43 & 0.49 & 51.63 \\
\textbf{Overpass} & 4.38 & 2.95 & 42.08 & 1.25 & 50.67 \\
\bottomrule
\end{tabular}}
\vspace{-3mm}
\end{table}

%% file: sections/5_conclusion.tex
 \section{Conclusion}
\label{sec:conclusion}
In this paper, we presented TRaIL-Odom, a tightly coupled Radar-IMU-LiDAR optimization framework based on continuous time B-splines for resilient state estimation across diverse environments. TRaIL-Odom modulates the contribution of radar Doppler constraints according to LiDAR geometric observability, reweighting Doppler residuals both per measurement and per scan.
Experimental results demonstrate robust performance in geometrically well- and ill-constrained scenes. In particular, enabling radar typically yields comparable or improved accuracy, showing that direction-dependent reweighting reinforces weakly observable directions without unnecessarily increasing radar influence along well-constrained directions. Ablation further confirms the benefit of the two weighting modules, reducing mean ATE and RTE by 86.0\% and 78.5\% over the fixed-weight baseline.
Future work will focus on integrating more complete covariance propagation and joint observability analysis of scene geometry, kinematic measurements, and sensor-specific uncertainty within the proposed adaptive weighting framework.